\documentclass[pdflatex,sn-basic]{sn-jnl}

\usepackage{subfig} %
\usepackage{array}  %
\usepackage{graphicx}%
\usepackage{multirow}%
\usepackage{amsmath,amssymb,amsfonts}%
\usepackage{amsthm}%
\usepackage{mathrsfs}%
\usepackage[title]{appendix}%
\usepackage{xcolor}%
\usepackage{textcomp}%
\usepackage{manyfoot}%
\usepackage{booktabs}%
\usepackage{algorithm}%
\usepackage{algorithmicx}%
\usepackage{algpseudocode}%
\usepackage{listings}%

\theoremstyle{thmstyleone}%
\theoremstyle{thmstyletwo}%

\theoremstyle{thmstylethree}%

\begin{document}

\title[Article Title]{ESNERA: Empirical and semantic named entity alignment for named entity dataset merging}


\author[1,2]{\fnm{Xiaobo} \sur{Zhang}}\email{zhangxiaobo@student.usm.my}

\author[2]{\fnm{Congqing} \sur{He}}\email{hecongqing@student.usm.my}

\author[1,2]{\fnm{Ying} \sur{He}}\email{heying11@student.usm.my}

\author[1]{\fnm{Jian} \sur{Peng}}\email{2007010@jxvc.jx.cn}

\author[1]{\fnm{Dajie} \sur{Fu}}\email{fdj0510@126.com}


\author*[2]{\fnm{Tien-Ping} \sur{Tan}}\email{tienping@usm.my}

\affil[1]{\orgdiv{School of Information Engineering}, \orgname{Jiangxi Vocational College of Finance \& Economics}, \orgaddress{ \city{Jiujiang}, \postcode{332000}, \state{Jiangxi}, \country{China}}}
\affil*[2]{\orgdiv{School of Computer Sciences}, \orgname{Universiti Sains Malaysia}, \orgaddress{\postcode{11700}, \state{Penang}, \country{Malaysia}}}


\abstract{Named Entity Recognition (NER) is a fundamental task in natural language processing. It remains a research hotspot due to its wide applicability across domains. Although recent advances in deep learning have significantly improved NER performance, they rely heavily on large, high-quality annotated datasets. However, building these datasets is expensive and time-consuming, posing a major bottleneck for further research. Current dataset merging approaches mainly focus on strategies like manual label mapping or constructing label graphs, which lack interpretability and scalability. To address this, we propose an automatic label alignment method based on label similarity. The method combines empirical and semantic similarities, using a greedy pairwise merging strategy to unify label spaces across different datasets. Experiments are conducted in two stages: first, merging three existing NER datasets into a unified corpus with minimal impact on NER performance; second, integrating this corpus with a small-scale, self-built dataset in the financial domain. The results show that our method enables effective dataset merging and enhances NER performance in the low-resource financial domain. This study presents an efficient, interpretable, and scalable solution for integrating multi-source NER corpora.}




\maketitle

\section{Introduction}\label{sec1}

Named Entity Recognition (NER) is a process to extract and classify named entities in text. NER plays a pivotal role in downstream applications such as information extraction, knowledge graph construction, question-answering systems, and etc\cite{1Aliod2006NamedER, 2Guo2009NamedER, 3Lample2016NeuralAF, 4Qu2023ASO}. With the advancement of transfer learning techniques, the utilization of pre-trained language models in training NER models has made significant progress in NER\cite{5Hu2024DeepLF, 6Abadeer2020AssessmentOD, 7chang2021chinese, 8mehta2021named, 9su2022roberta, 10kaur2024bert}.

The availability of high-quality NER corpora is crucial for training high-performance NER models. However, many NER datasets have the following limitation. Firstly, most NER datasets are of medium or small scale and cover limited domains. Thus, it may fail to satisfy the practical application demands of diverse tasks and industries. For example, OntoNotes 5.0\cite{11weischedel2013ontonotes} primarily focuses on news texts, while CLUENER2020\cite{12xu2020cluener2020} spans multiple domains but lacks sufficient scale. Secondly, there may be variations in the definition of the name entity, annotation granularity, and annotation norms across different datasets. For instance, the definition of label \texttt{GPE} in OntoNotes 5.0 differs from the label \texttt{address} in CLUENER2020, leading to inconsistencies in label schemas. The issue of label inconsistency makes it arduous for existing models to directly utilize data from diverse sources for joint training or transfer learning. Such disparities can lead to conflicts during model training, thereby reducing the performance of NER models. Previous studies address the label inconsistency through building a domain knowledge-based label graph\cite{13zhao2021joining} and pseudo-labeling\cite{14Arazo2019PseudoLabelingAC}; however, these methods typically suffer from poor interpretability and limited expandability in fusing label systems. This study aims to address these limitations through the following objectives: \\
(1) To develop a method for aligning named entity labels in different datasets by analyzing their similarity.\\
(2) To construct a unified large-scale NER corpus through progressive dataset merging based on the label alignment. \\
(3) To evaluate the effectiveness of the proposed label merging method on NER performance in general domains.\\
(4) To further verify its cross-domain transferability and robustness in low-resource scenarios using a small-scale financial dataset (FinReportNER).

To achieve this, we argue that it is essential to explore set-theoretic relationships between named entities, such as equivalence, subset/superset, partial overlap, and disjointness, and to develop a named entity alignment strategy that unifies consistent label pairs across datasets by identifying them both semantically and empirically. This approach facilitates scalable and reliable cross-domain datasets merging. We introduce two complementary similarity measures: Empirical similarity measures the proportion of entity overlap between different datasets, which can reveal commonalities in annotation standards and label granularity. Semantic similarity, computed from contextual embeddings, measures the semantic proximity between label representations. These two types of similarity metrics possess complementary advantages. We comprehensively consider them through a linear interpolation fusion approach, formulating a label merging strategy that is highly interpretable and practical.

We perform our study in Chinese NER. Our experiment was conducted on three mainstream Chinese NER datasets: OntoNotes5.0, CLUENER2020, and BosonNER\cite{15BosonNLP}. Hereafter, we refer to them as \texttt{OntoNotes, CLUENER, and BosonNER}, respectively. The main contributions of this paper are summarized as follows:\\
(1) We propose an automatic named entity alignment method based on empirical and semantic similarity.\\
(2) We develop a scalable merging framework using a greedy pairwise alignment strategy and grid search to maximize named entity merging while minimizing performance degradation.\\
(3) We prove the effectiveness of the proposed approach by merging three NER datasets for training a NER.\\
(4) We validate the approach on a small financial NER dataset (FinReportNER), demonstrating its effectiveness and cross-domain adaptability.\\

\section{Related Works}\label{sec2}

\subsection{NER}\label{subsec2_1}

With the development of deep learning technology, NER methods have evolved from rule-based and statistical models to deep neural architectures. \textbf{Rule-based methods} rely on rules constructed by experts and regular expressions. Although these methods do not require labeled data and have good interpretability, they are subject to domain limitations due to the high maintenance costs associated with rule reconstruction\cite{16Rau1991ExtractingCN}. \textbf{Statistical machine learning-based methods} such as Hidden Markov Models\cite{17Zhou2002NamedER} and Conditional Random Fields\cite{18Lafferty2001ConditionalRF} solve the above problems by using probabilistic frameworks to learn dependencies between tokens and labels from annotated corpora. These models usually rely on carefully designed features, such as tf-idf\cite{19Ingle2017PredictiveMF}, syntactic, lexical, or morphological features, which may not be optimal. Although statistical models have good robustness and probabilistic interpretability, their reliance on manual features and difficulty in modeling long-distance dependencies limit their performance in complex language scenarios. \textbf{Deep learning-based methods} such as Convolutional Neural Network (CNN)\cite{20Ma2016EndtoendSL, 21Gui2019CNNBasedCN} and Recurrent Neural Network (RNN)\cite{22Cho2014LearningPR, 23Wu2017ClinicalNE} alleviate the reliance on manual features in traditional approaches by automatically learning representations from raw text. These models can capture complex nonlinear relationships and long-range dependencies, demonstrating significant performance improvements in multiple NER tasks. Their end-to-end training mechanism enhances the scalability and cross-domain adaptability of the methods.

The current mainstream NER model architectures are: 1) Encoder-type models: such as BERT-CRF, which utilize the pre-trained BERT encoder to extract contextual features and combine CRF for sequence labeling\cite{24Kaur2024BertNerAT, 25Chandra2024AviationBERTNERNE}, demonstrating high efficiency in multi-domain NER; 2) Encoder-Decoder-type models: such as T5\cite{28JMLR:v21:20-074} and BART\cite{27Lewis2019BARTDS}, which treat NER as a sequence generation task, with the encoder extracting features and the decoder generating entity labels \cite{26Ni2021SentenceT5SS}, suitable for complex annotation scenarios; 3) Decoder-only models: such as ChatGPT\cite{29Laskar2023ASS}, which generate entity labels through prompt learning\cite{30shen-etal-2023-promptner}, adapting well to zero-shot or low resource tasks. Table \ref{tab1} shows their mechanisms, advantages, limitations, and performance characteristics.

\begin{table}
\caption{NER Model Comparison}\label{tab1}%
    \begin{tabular}{>{\centering\arraybackslash}m{0.14\textwidth} 
                    >{\centering\arraybackslash}m{0.24\textwidth} 
                    >{\centering\arraybackslash}m{0.20\textwidth} 
                    >{\centering\arraybackslash}m{0.22\textwidth}}
    \toprule
    Model Type & Representative Models  & Advantages &  Limitations\\
    \midrule
    Encoder-only    & BERT\cite{31devlin2019bert}, RoBERTa\cite{32liu2019roberta}, BERT-BiLSTM-CRF\cite{33Dai2019NERBERTCRF}  & Stable performance, global deciding with CRF, low training cost  & Limited flexibility in nested entity extraction  \\
    Encoder-Decoder    & T5, BART   & Flexible format, handles complex annotations  & Higher computation, formatting-sensitive  \\
    Decoder-only    & GPT family(ChatGPT)\cite{34brown2020language}   & Zero-shot, Few-shot learning, language generation flexibility  & Exhibit prompt sensitivity, performance is lower than the encoder-only model in standard NER tasks  \\
    \botrule
    \end{tabular}
\end{table}

\subsection{NER Datasets}\label{subsec2_2}

Over the years, numerous NER datasets have been constructed using news, social media, and financial content\cite{35LI2020103435, 36ding-etal-2021-nerd, 37loukas-etal-2022-finer}. However, the diversity of these datasets presents important challenges in multi-dataset NER tasks, especially when merging datasets to construct a unified training corpus. Datasets exhibit notable differences in domain and contextual style. 
For instance, CoNLL-2003\cite{38CoNLL-2003} consists of English news texts, making it suitable for general-domain NER tasks. MSRA-NER\cite{39levow2006third} covers Chinese news texts with a formal tone. OntoNotes focuses on the field of news and includes trilingual corpora in English, Chinese, and Arabic. This design was promoted by the Linguistic Data Consortium (LDC) to support cross-lingual natural language processing research. Its data is sourced from newswire, broadcast news, and web data \cite{11weischedel2013ontonotes}. CLUENER spans multiple domains, such as encyclopedia entries, news, and question-answering texts, offering diverse contexts; BosonNER, derived from social media, features a colloquial style, published by bosonNLP\cite{15BosonNLP}; Zhang et al.\cite{41zhang2021named} developed a financial NER dataset based on enterprise annual reports for enterprise evaluation systems. Finer-139\cite{37loukas-etal-2022-finer} dataset was proposed in the financial domain, based on XBRL annotations. Shah et al.\cite{42Shah2023} developed a high-quality corpus focused on financial entity recognition. 

These datasets have several challenges when attempting integration. Firstly, domain differences lead to variations in entity distributions and contextual styles, such as the colloquial entities in BosonNER (e.g., ``@user'') versus the formal entities in financial datasets (e.g., ``company names''). Secondly, semantic divergence arises due to domain-specific interpretations. For example, in CLUE,``apple'' refers to \texttt{movie}, while in Wang et al. (2021)'s financial data, ``apple'' refers to \texttt{ORGANIZATION} (Apple Inc.). Similarly, ``Beijing Haidian District'' may be labeled as \texttt{GPE} in OntoNotes, while ``New York Stock Exchange'' is tagged as \texttt{LOCATION} in Finer-139, even though both refer to geopolitical entities. Additionally, variations in dataset size and label schemas further complicate the integration process. Some datasets are large-scale (e.g., MSRA-NER and OntoNotes). In contrast, others, such as BosonNER, have fewer samples, which may lead to imbalanced distributions in the merged corpus and affect its representativeness. Label schemas also differ in complexity: CoNLL-2003 and MSRA-NER have simpler schemas, while OntoNotes and CLUE feature more detailed labels, and financial datasets include domain-specific categories. These differences result in label inconsistencies during the merging process. For example, ``Harry Potter'' being labeled as \texttt{movie}, \texttt{person}, or \texttt{book} across datasets, or ``Beijing'' annotated as \texttt{GPE} in some datasets but \texttt{location} in others. Moreover, partially overlapping labels (e.g., \texttt{scene} in CLUENER and \texttt{GPE} in OntoNotes) may introduce noise due to stylistic or domain discrepancies, and smaller datasets risk being overshadowed by larger ones, reducing corpus diversity. These challenges highlight the complexity of label alignment and integration in multi-dataset NER. Table \ref{tab2} shows a comparative overview of the NER Datasets mentioned above.

\begin{table}
\caption{Comparative Overview of Public NER Datasets}\label{tab2}%
\begin{tabular}{>{\centering\arraybackslash}m{0.10\textwidth} 
                >{\centering\arraybackslash}m{0.10\textwidth} 
                >{\centering\arraybackslash}m{0.15\textwidth}
                >{\centering\arraybackslash}m{0.15\textwidth} 
                >{\centering\arraybackslash}m{0.10\textwidth}
                >{\centering\arraybackslash}m{0.20\textwidth}}

\toprule
Dataset & Domain & Language & Size(Train/ Dev/ Test) & \#Labels & Notes/Features \\
\midrule
MSRA-NER & Chinese & News & ~46K / — / ~4K & 3 & Only PER/LOC/ORG; widely used in Chinese NER benchmarks  \\
CoNLL-2003 & English & News(Reuters) & ~15K / ~3.5K / ~3.5K & 4 & Classic benchmark; BIO format; limited to 4 entity types \\
OntoNotes 5.0 & Chinese, English, Arbic & Multi-domain(news, talk, etc.) & ~120K / ~16K / ~24K & 18+ & Rich tag set; covers multiple languages and genres \\
CLUENER 2020 & Chinese & Online comments, news & ~10K / ~1.3K / ~1.3K & 10 & Fine-grained tags like book, game, movie; crowd-annotated \\
BosonNLP NER & Chinese & Social media, news & ~2K / ~0.3K / ~0.3K & 8 & Small-scale \\
\botrule
\end{tabular}
\end{table}

\subsection{Label Alignment and Dataset Merging Methods}\label{subsec2_3}

Studies have proposed named entity alignment methods to merge datasets, which can be divided into three categories: manual mapping, constructing label graphs, and pseudo-labeling. In early studies, manual mapping methods align labels by defining mapping rules by experts. However, this approach relies heavily on expert participation, and it is costly and has limited scalability, making it difficult to adapt to new datasets. Zhao et al.\cite{13zhao2021joining} unified multiple datasets from the same domain by constructing a knowledge-driven label graph. This method leverages existing classification structures from pet websites to establish mapping relationships between labels with different levels of detail or hierarchy. The label graph combines original label nodes from different datasets (such as fine-grained cat and dog breeds) with augmented nodes (such as color or hair features) to create a data merging pathway. This method relies on the existing classification structure of pet websites to build mapping relationships between labels. It is mainly applicable to datasets from the same source, such as those for cats and dogs. Although this method significantly reduces costs compared to manual label mapping, its scalability on heterogeneous datasets is limited. Pseudo-labeling methods attempt to solve the label inconsistency problem through cross-dataset training, such as training a model on a source dataset and generating pseudo-labels on a target dataset \cite{43lee2013pseudo, 44arazo2020pseudo}. However, these methods may introduce noise and lack interpretability, making it difficult to ensure the accuracy of alignment. Additionally, some methods align labels by leveraging the semantic embeddings of the labels themselves, such as calculating the embeddings of labels using BERT\cite{45ma-etal-2022-label}. Although these methods capture the semantic relationships between labels to some extent, they ignore the context semantics of entities and are susceptible to annotation noise, with limited effectiveness in handling partially overlapping relationships (such as ``scene'' and ``GPE''). The above methods have provided beneficial attempts for label alignment and dataset merging, but there is still room for improvement.

\section{Methodology}\label{sec3}

As mentioned in Section \ref{sec2}, the inconsistency in naming entities across datasets presents a major challenge for joint training in the NER task. Existing methods have significant limitations: manual mapping and label graph construction depend on expert knowledge and domain knowledge, which are costly and poorly scalable; approaches that only utilize label semantics overlook the commonalities in dataset annotation practices and the contextual semantics of entities, making adaptation to differences in annotation distributions difficult. Additionally, current methods often do not fully account for the set relationships among labels (such as equivalence, subset/superset, partial overlap, and disjointness), leading to unsystematic alignment and poor integration of different datasets. To address these gaps, we propose ESNERA, a named entity alignment method that systematically models label relationships and supports scalable, interpretable, and automated merging of multi-source NER datasets.
The core idea of ESNERA is to identify alignment relationships between labels from different datasets through similarity-based estimation, rather than explicitly determining their set-theoretic types. We conceptually define four types of label relations: (1) equivalence, (2) subset/superset, (3) partial overlap, and (4) disjointness. Different from previous works, our method does not require prior knowledge to categorize them. Instead, it calculates a combined similarity score $S_{merge}(L_s, L_t)$ between each pair of source and target labels. If the score surpasses a threshold $\tau$, the two labels are considered semantically similar and are merged. In practice:
\begin{itemize}
  \item High S\_merge(L\_s, L\_t) scores often correspond to equivalence (e.g., \texttt{name} in CLUE and \texttt{PERSON} in OntoNotes);
  \item Moderate S\_merge(L\_s, L\_t) scores may reflect partial overlap or subset/superset relations (e.g., \texttt{company\_name} in BosonNER and \texttt{ORG} in OntoNotes);
  \item Low S\_merge(L\_s, L\_t) scores typically indicate disjoint labels (e.g., \texttt{location} in BosonNER and \texttt{FAC} in OntoNotes).
\end{itemize}

\begin{figure}
    \centering
    \includegraphics[width=0.8 \linewidth]{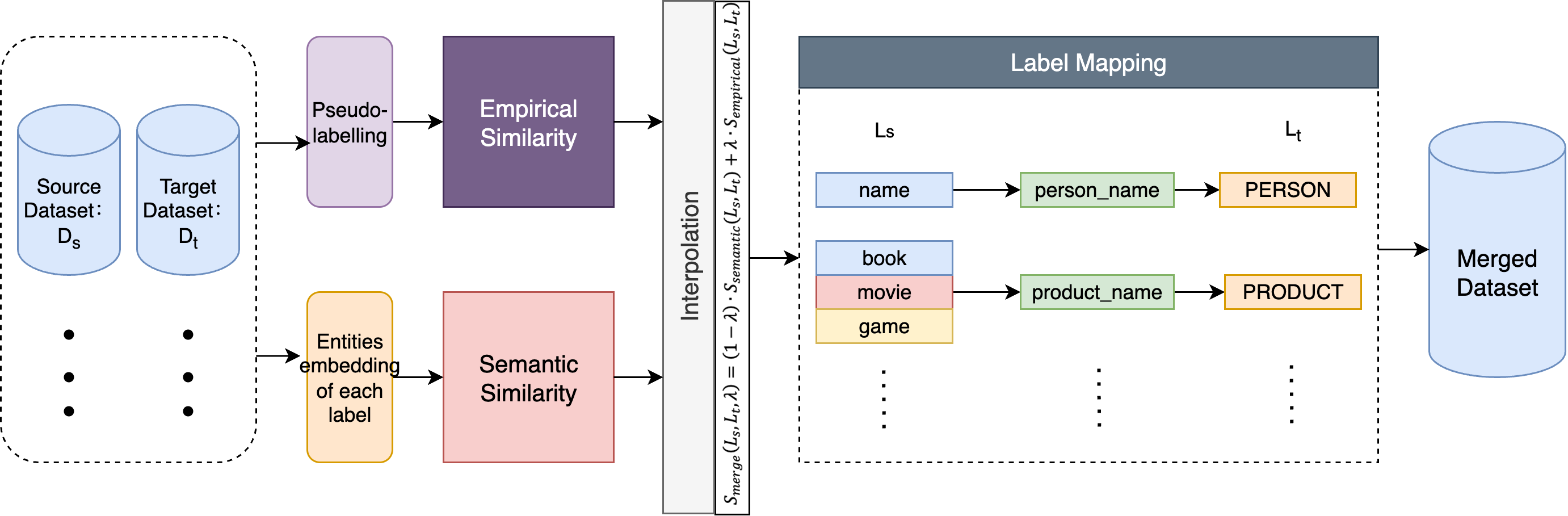}
    \caption{The overall structure of the proposed framework}
    \label{fig:overall_framework}
\end{figure}

This integrated similarity-based methodology eliminates the necessity for manual classification of label relations while maintaining effective management of diverse merging scenarios. The proposed method comprises four steps: 1) label similarity computation, quantifying empirical and semantic similarities between labels; 2) grid search for optimal merging parameters, determining similarity thresholds and weights; 3) automatic annotation of missing labels, using contextual information to annotate unlabeled entities; and 4) label merging and corpus integration, producing a unified training corpus. An overview of the pipeline is shown in Figure \ref{fig:overall_framework}. 

\subsection{Label Similarity Computation}\label{subsec3_1}

We calculate pairwise similarities between labels from n dataset pairs to determine which labels from different datasets are semantically and statistically aligned. For each label pair $(L_s,L_t)$, $L_s$ is source dataset label, $L_t$ is target dataset label. We use two complementary similarity metrics: empirical similarity and semantic similarity.

\subsubsection{Empirical similarity}\label{subsubsec3_1_1}
\begin{figure}
    \centering
    \includegraphics[width=0.8 \linewidth]{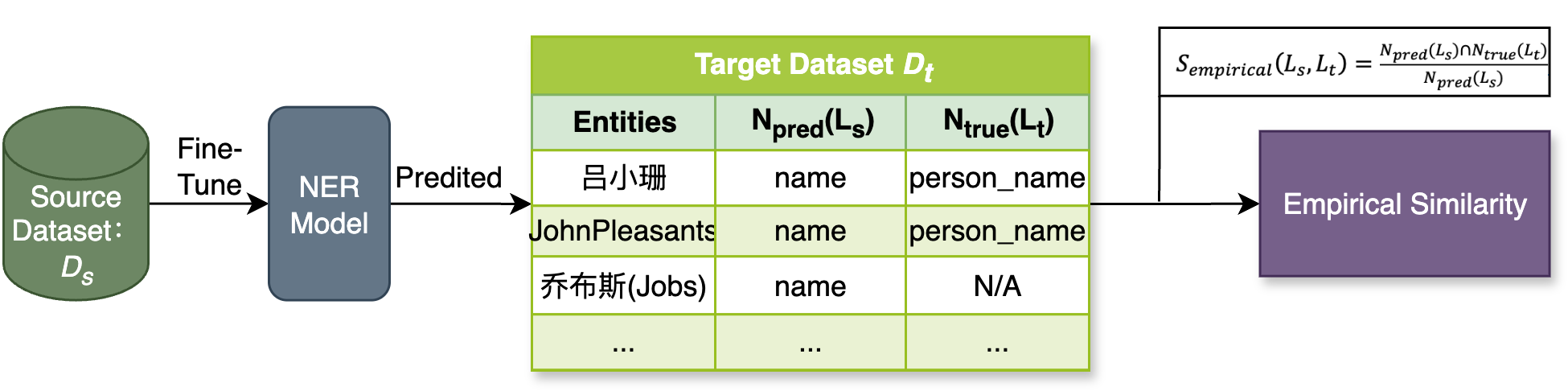}
    \caption{The structure of empirical similarity}
    \label{fig:2}
\end{figure}

The empirical similarity $S_{empirical}(L_s,L_t)$ is used to measure the matching of the name entity pair $(L_s, L_t)$ in the pseudo-labelling task between the source dataset $D_s$ and the target dataset $D_t$, and its calculation is based on the prediction results of the NER model. The structure of the empirical similarity module is shown in Figure \ref{fig:2}. Specifically, we first train an NER model on the source dataset $D_s$, and then use this model to evaluate on the training set of the target dataset $D_t$ to identify entities and compare the matching degree of the predicted labels with the true labels. The calculation process is as follows: Suppose $L_s$ is the label in the source dataset (such as \texttt{address} in CLUE), and $L_t$ is the label in the target dataset (such as \texttt{GPE} in OntoNotes), we use the NER model trained on $D_s$ to predict all entities on the training set of $D_t$. For all entities in $D_t$ that are truly labeled as $L_t$, the model may predict several entities as $L_s$. The empirical similarity is defined as the proportion of entities truly labeled as $L_t$ among the entities predicted as $L_s$ by the model. The formula is as follows:

\begin{equation}
S_{empirical}\left(L_s, L_t\right)=\frac{N_{pred}(L_s)\cap N_{true}(L_t)}{N_{pred}(L_s)}\label{eq1}
\end{equation}

Herein, $N_{pred}(L_s)$ denotes the number of entities predicted as $L_s$ by the model on $D_t$, and $N_{pred}(L_s)\cap N_{true}(L_t)$ represents the number of these predicted entities that are truly labeled as $L_t$. For example, if the model predicts 100 entities as ``address" on the training set of $D_t$ (i.e., $N_{pred}(address)=100$), and among them, 70 entities have the true label of \texttt{GPE} (i.e., $N_{pred}(address)\cap N_{true}(GPE) =70$), then $S_{empirical}(address, GPE)=70/100=0.7$, or 70\%. This metric reflects the consistency of label prediction of the model trained on $D_s$ when applied to $D_t$ effectively capturing the commonalities between the two labels in the annotation practice. 

In particular, empirical similarity exhibits asymmetry and sensitivity to direction, which means that $S_{empirical}(L_s,L_t)\ \neq\ S_{empirical}(L_t,L_s)$. This is due to differences in label schemas, data domains, and annotation norms between $D_s$ and $D_t$. When the direction is reversed (i.e., training the model with $D_t$ and making predictions on $D_s$), the prediction results are different. For example, a model trained on OntoNotes may frequently predict \texttt{GPE} as \texttt{address}, but the converse may not necessarily be true. Consequently, it is necessary to calculate the empirical similarity in both directions. This also gives rise to the problem of choosing the merging path when integrating multiple datasets. This issue will be further explored in Section \ref{subsec4_2} to devise an effective merging path selection strategy.

\subsubsection{Semantic Similarity}\label{subsubsec3_1_2}

While empirical similarity reflects annotation behavior, it may fail if limited data is annotated. To address this limitation, we compute the semantic similarity $S_{semantic}(L_s,L_t)$ by comparing the contextual word embeddings of entities associated with each label. Word embeddings is a technique that maps words or phrases into a low-dimensional real vector space, enabling the capture of the semantic and contextual information of words. Traditional word embedding methods, such as Word2Vec\cite{46levy2014linguistic} and GloVe \cite{47pennington2014glove}, generate static embeddings. In this case, the embedding vector of each word remains fixed and cannot adapt to contextual variations. Conversely, pre-trained language models based on Transformer\cite{48vaswani2017attention}, like BERT\cite{31devlin2019bert}, generate dynamic embeddings through context awareness. These models can dynamically adjust the vector representation of a word according to its context, thus more accurately capturing semantic information. In this study, we employ the BERT-based model, specifically the Chinese pre-trained model based on the Whole Word Masking strategy. This model outperforms the original BERT in terms of Chinese word segmentation consistency and context modeling. It is particularly well-suited for handling long word structures and complex contexts in the Chinese language. The calculation of semantic similarity is divided into the following three steps: extracting the word embeddings of named entities, centralizing and normalizing the word embeddings, and calculating the cosine similarity. The structure of the semantic similarity module is shown in Figure \ref{fig:3}. The specific processes are as follows:

\begin{figure}
    \centering
    \includegraphics[width=0.8\linewidth]{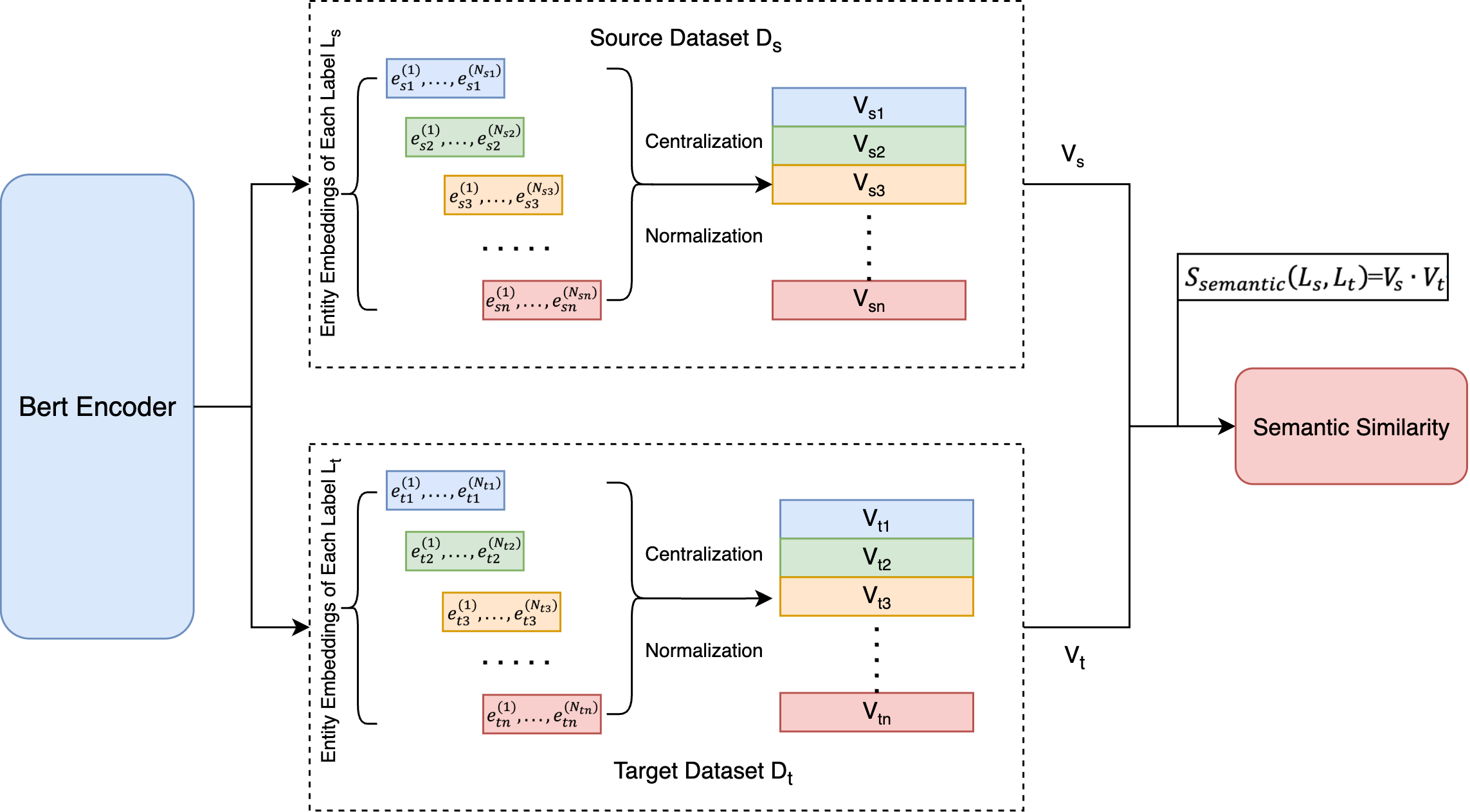}
    \caption{The structure of semantic similarity}
    \label{fig:3}
\end{figure}

\begin{enumerate}
    \item Entity Embeddings Extraction \\
    To compute the semantic similarity of the label pair $(L_s,L_t)$, we initially extract all entities labeled as $L_s$ from the source dataset $D_s$ and those labeled as $L_t$ from the target dataset $D_t$. For each entity (e.g., \texttt{movie}, \texttt{organization}), we feed the sentence in which it resides into the BERT model to obtain its contextual embedding. Subsequently, we identify the tokens that correspond to the entity span based on the character offsets and calculate the average of their embeddings. This mean vector represents the contextual semantics of the entity. This procedure ensures that the entity embeddings can reflect their contextual semantics. As a result, it provides a reliable basis for subsequent similarity computations.
    
    \item Centralization and Normalization \\
    After extracting the entity embeddings, we perform centralization and normalization on the embedding vectors of all entities to eliminate potential offsets and dimensional differences, ensuring the accuracy of cosine similarity calculations. \\
    \textbf{Centralization processing}: The centralization eliminates the global offset of the embedded vectors in the semantic space, making the embedded distributions of different labels more comparable\cite{49fuhl2021weight, 50aboagye2022normalization}. Suppose the entity set of labels $L_s$ contains $N_s$ entities, and their corresponding embedded vectors are $\mathcal{E}_s = \{e_{s}^{(1)}, e_{s}^{(2)}, \dots, e_{s}^{(N_s)}\}$. We first calculate the mean vector $\mu_s$ of these embedded vectors:
    \begin{equation}
        \mu_s = \frac{1}{N_s} \sum_{i=1}^{N_s} e_{s}^{(i)} \label{eq2}
    \end{equation}
    Subsequently, each embedded vector is centered by subtracting the mean vector:
    \begin{equation}
        \tilde{e}_{s}^{(i)} = e_{s}^{(i)} - \mu_s, \quad i = 1, 2, \dots, N_s \label{eq3}
    \end{equation}
    \textbf{Normalization processing}: The normalization eliminates the dimensional differences of the embedding vectors, ensuring that the cosine similarity only reflects the directional differences between vectors and is not affected by vector length\cite{51arora2019character, 52obidallah2022multi}. The centralized embedding vectors $\tilde{e}_{s}^{(i)}$ may have different lengths, which can affect the calculation of cosine similarity. To address this issue, we normalize each centralized embedding vector to have a length of 1:
    \begin{equation}
        \hat{e}_{s}^{(i)} = \frac{\tilde{e}_{s}^{(i)}}{\left\| \tilde{e}_{s}^{(i)} \right\|}, \quad i = 1, 2, \dots, N_s \label{eq4}
    \end{equation}
    where $\left\| \tilde{e}_{s}^{(i)} \right\|$ represents the L2 norm of the vector $\tilde{e}_{s}^{(i)}$, that is, $\sqrt{ \sum_{j=1}^{768} \left( \tilde{e}_{s}^{(i)} \right)^2 }$. 
    
    \item Cosine Similarity \\
    After completing the centralization and normalization processing, we calculate the semantic similarity of the labels $L_s$ and $L_t$ using cosine similarity\cite{53xia2015learning, 54lahitani2016cosine}. For $L_s$, we take the average of the embedding vectors of all its entities after centralization and normalization to obtain the average embedding vector $V_s$ of the label:
    \begin{equation}
        V_s = \frac{1}{N_s} \sum_{i=1}^{N_s} \hat{e}_{s}^{(i)} \label{eq5}
    \end{equation}
    Similarly, for $L_t$, we compute its mean embedding vector $V_t$. Then the semantic similarity is determined using the cosine similarity equation:

    \begin{equation}
       S_{\text{semantic}}(L_s, L_t) = \cos(V_s, V_t) = \frac{V_s \cdot V_t}{\left\| V_s \right\| \left\| V_t \right\|} \label{eq6}
    \end{equation}

    Since $V_s$ and $V_t$ have already been normalized, simplifying the equation to:

    \begin{equation}
       S_{\text{semantic}}(L_s, L_t) = V_s \cdot V_t \label{eq7}
    \end{equation}

    The cosine similarity value lies within the range of [-1, 1]. A value closer to 1 indicates a higher degree of semantic similarity between the two labels in the semantic space. For instance, if the average embedding vectors of \texttt{address} and \texttt{location} are proximate in the semantic space, the resulting value of $S_{semantic}(address, location)$ will be closer to 1, thereby signifying a robust semantic correlation between the two labels.
    
\end{enumerate}

\subsubsection{Merged Similarity}\label{subsubsec3_1_3}
Empirical similarity and semantic similarity are complementary: empirical similarity reflects the annotation norms of each dataset but may be influenced by data distribution; semantic similarity captures the semantic proximity between labels but is sensitive to annotation noise. To fully leverage the advantages of both, we calculate the combined similarity $S_{merge}(L_s,L_t,\lambda)$ through linear interpolation:
    
    \begin{equation}
       S_{\text{merge}}(L_s, L_t, \lambda) = (1 - \lambda) \cdot S_{\text{semantic}}(L_s, L_t) + \lambda \cdot S_{\text{empirical}}(L_s, L_t) \label{eq8}
    \end{equation}
where $\lambda \in [0, 1]$ is a tuning parameter used to balance the contributions of semantic similarity and empirical similarity. When the value of $\lambda$ is low, the model leans more towards semantic similarity, emphasizing the semantic closeness between labels; when the value of $\lambda$ is high, the model pays more attention to empirical similarity, highlighting the commonalities in annotations. This study determined the optimal value of $\lambda$ through experiments (see Section \ref{subsec4_3}) to achieve the best balance between label merging and NER performance.

\subsection{Label Merging Paths and Strategies}\label{subsec3_2}

Due to the heterogeneity of label definitions among different datasets, choosing an appropriate merging path is paramount for label alignment and model performance accuracy. To circumvent the combinatorial explosion resulting from a one-time global merge, this paper proposes a unidirectional similarity greedy merging strategy, achieving efficient and stable label fusion via pairwise dataset alignment and label mapping prioritized by maximum empirical similarity.

\subsubsection{Pairwise Merging Strategy}\label{subsubsec3_2_1}

The method applies pairwise dataset merging, aligning the named entities of only two datasets in each round to generate an intermediate dataset. In the subsequent round, this intermediate dataset serves as the basis for alignment with the remaining unmerged datasets until all datasets are integrated. This strategy effectively mitigates the complexity of the merging process and reduces the deviation of label semantics.

\subsubsection{Unidirectional Similarity Greedy Merging Strategy}\label{subsubsec3_2_2}

When the number of datasets to merge exceeds three, the number of merging paths grows exponentially, making it infeasible to enumerate all combinations. Hence, this paper proposes a unidirectional similarity greedy merging strategy, based on the concept of the greedy algorithm\cite{57zhang2000greedy}. Each round selects the pair with the highest unidirectional empirical similarity to construct the globally optimal path. The specific process is as follows:
\begin{enumerate}
    \item Calculate unidirectional empirical similarity: For each pair of datasets $(D_s,D_t)$, we calculate the unidirectional empirical similarity $S_{\text{empirical}}(L_s,L_t)$ for all label pairs $(L_s \in D_s, L_t \in D_t)$, and get the sum of empirical similarity $\sum_{(D_s, D_t)} S_{\text{empirical}}(L_s, L_t)$. The influence of invalid and missing values (NaN) is excluded to ensure robustness.
    \item Initialize the merging path: To generate the first intermediate dataset, select the two datasets with the highest sum of empirical similarity calculated by the last step as the initial merging pair.
    \item Iterative optimal selection: In each round of merging, from the unmerged datasets, select the one with the highest unidirectional empirical similarity to the current intermediate dataset for the next round of merging and update the intermediate dataset.
    \item Termination condition: Repeat the above steps until all datasets are incorporated into the merging path. 
\end{enumerate}
This strategy maximizes the cumulative sum of unidirectional empirical similarities, prioritizing the merging pairs with the most similar label distributions. This is equivalent to selecting the cumulative sum of high-similarity paths in the empirical similarity matrix. The exclusion of NaN values ensures the calculation's stability and objectivity, rendering this method efficient and scalable in multi-dataset scenarios.

\subsubsection{Label Mapping Strategy}\label{subsubsec3_2_3}

During the label alignment process, some source labels may have multiple candidate target labels. For example, the label \texttt{time} in BosonNER may correspond to both the \texttt{DATE} (similarity = 0.77) and \texttt{TIME} (similarity = 0.40) labels in OntoNotes. However, assigning a single entity to multiple target labels is neither practical nor desirable, as it would introduce ambiguity and redundancy in the merged dataset. To address this, we adopt a maximum similarity priority strategy, in which each source label is aligned to only one target label, the one with the highest empirical similarity score. This decision ensures a clear and deterministic mapping, reduces alignment noise, and enhances the overall robustness and interpretability of the label alignment process.

\subsection{Grid Search for Parameter Optimization}\label{subsec3_3}
This paper uses the grid search approach to optimize the parameters during the label merging process. The system systematically traverses the weighting coefficient $\lambda$ and the merging threshold $\tau$ to maximize the quantity of merged labels while ensuring minimal variation in the F1 score. The grid search is founded on the comprehensive similarity $S_{merge}$ defined in Section \ref{subsec3_1}. It integrates the unidirectional similarity greedy merging strategy from Section \ref{subsec3_2} to furnish a robust label space for the NER task of multi-source datasets.

\subsubsection{Evaluation metrics}\label{subsubsec3_3_1}
\begin{itemize}
    \item The number of merged labels: Defined as the total number of different labels mapped to the same target label. For instance, if \texttt{company},\texttt{organization}, and \texttt{government} are merged into \texttt{ORG}, it is counted as three merged labels. This indicator reflects the coverage and diversity of the label alignment.
    \item Data row increment: Defined as the increment of the sample count (data rows) of the merged dataset in relation to the original dataset, calculated as:
    \item F1 Score: The F1 score is adopted to measure the NER performance of the model on the test set. The F1 score is based on precision and recall and is defined as follows\cite{58Li2020, 59Popovski2020}:
    \begin{equation}
        \text{Precision} = \frac{\text{TP}}{\text{TP} + \text{FP}} \label{eq9}
    \end{equation}
    \begin{equation}
        \text{Recall} = \frac{\text{TP}}{\text{TP} + \text{FN}} \label{eq10}
    \end{equation}
    \begin{equation}
        F_1 = \frac{2 \times \text{Precision} \times \text{Recall}}{\text{Precision} + \text{Recall}} \label{eq11}
    \end{equation}
    where TP (True Positives) represents the number of correctly predicted entities, FP (False Positives) represents the number of wrongly predicted entities, and FN (False Negatives) represents the number of missed entities.
    \end{itemize}
    
In this study, we also report the micro-averaged F1 score, which reflects the contributions of all labels by computing the global counts of TP, FP, and FN. The micro-averaged F1 score is defined as:

\begin{equation}
    \text{Micro-}F_1 = \frac{2 \times \sum \text{TP}}{2 \times \sum \text{TP} + \sum \text{FP} + \sum \text{FN}} \label{eq:microf1}
\end{equation}

\subsubsection{Grid Search Procedure}\label{subsubsec3_3_2}

The grid search optimizes the parameters $\lambda$ and $\tau$ through the following steps. Here, $\tau$ represents the threshold for the comprehensive similarity $S_{merge}$, determining whether a label pair is sufficiently similar for merging.
\begin{enumerate}
    \item \textbf{Parameter Range:} Based on the pre-researches in multi-source similarity fusion tasks\cite{55lin-2004-rouge, 56Reimers2019}, we set $\lambda \in \{0.3, 0.4, 0.5, 0.6, 0.7\}$ and $\tau \in \{0.1, 0.2, 0.3, 0.4, 0.5, 0.6, 0.7, 0.8, 0.9\}$, covering typical trade-off configurations between empirical and semantic similarity, as well as the effective interval for the merging threshold.
    \item \textbf{Label Merging:} For each parameter combination $(\lambda, \tau)$, calculate the label similarity based on $S_{\text{merge}}$ in Section \ref{subsec3_1}, execute the greedy merging strategy described in Section \ref{subsec3_2}, generate the merged label set and dataset, and record the number of merged labels.
    \item \textbf{Performance Evaluation}: Fine-tune the NER model using the merged dataset and assess the F1 score on the test set of the target task. Compare it with the baseline model without merging to guarantee that the F1 score change is within an acceptable range (fluctuation less than 2\%).
    \item \textbf{Parameter Selection}: If the F1 score is comparable to the baseline, select the parameter combination with the maximum number of merged labels as the optimal configuration.
\end{enumerate}

\subsubsection{Data Preprocessing}\label{subsubsec3_3_3}

To enhance the efficiency of the grid search, when calculating $s_{empirical}$, pre-filter low-frequency labels (occurring less than 5 times) and invalid values (NaN) to ensure the stability of the similarity calculation.

\subsection{Label Augmentation}\label{subsec3_4}

During the integration of multiple datasets, specific labels present in the source dataset may be missing from the target dataset. For example, numerical entity types like \texttt{PERCENT} and \texttt{TIME} exist in the OntoNotes dataset but do not have direct equivalents in the CLUE dataset. Experimental results show that without addressing this issue, the merged model cannot recognize numerical entity types in NER tasks on different datasets, which greatly reduces downstream performance. To ensure the label space in the merged dataset is complete and to support robust multi-source NER training, we use pseudo-labeling to fill in missing labels.
Specifically, for labels not present in the target dataset, we utilize a pretrained NER model to generate pseudo-labels. For instance, to handle the absence of the \texttt{CARDINAL} entity type in the CLUE dataset, we fine-tune a BERT-CRF model on the OntoNotes dataset and apply it to the CLUE corpus to predict the relevant entities. This method works particularly well for entity types with unique contextual patterns, such as numerical or temporal entities, where the model can accurately infer labels based on learned representations.
The proposed label augmentation module greatly improves the label space coverage in the merged dataset without sacrificing label accuracy. This provides a strong foundation for multi-source NER training, helping the model to generalize effectively across different datasets.

\subsection{Baseline Methods}\label{subsec3_5}

To assess the validity of the proposed approach, this paper devises the following two baseline methods for comparison with the automatic label merging strategy:

\textbf{Baseline 1 Independent Training}: No label merging is carried out. The BERT-CRF model is independently trained on the CLUENER, BosonNER, and OntoNotes datasets, respectively. Each model employs the same architecture and training parameters (refer to Section \ref{subsec4_1} for details) and conducts entity recognition merely based on the label system of the individual dataset. This approach represents the original performance without merging and is applicable for evaluating the model's generalization ability enhancement due to label merging.

\textbf{Baseline 2 Manual Label Merging}: Through manual screening of the datasets, a label mapping table is established for label merging. The process involves domain experts analyzing the label systems of CLUENER, BosonNER, and OntoNotes, and manually developing one-to-one or one-to-many label mapping rules based on label definitions and semantic relationships (for example, mapping \texttt{address} in CLUENER to \texttt{GPE} or \texttt{FAC} in OntoNotes). Then, the label systems are aligned using these rules, the datasets are merged, and the BERT-CRF model is trained. This approach demonstrates traditional manual label alignment performance and serves as a comparison point for assessing the efficiency and accuracy of the automatic merging strategy.
The experimental settings of the baseline methods, the number of merged labels, the increment of data rows, and the performance results are elaborated in Section \ref{sec4}.

\section{Experiments and Results}\label{sec4}

\subsection{Experiment Setup}\label{subsec4_1}

\textbf{Datasets: }
We conduct the study on Chinese NER using three representative datasets: the Chinese portion of OntoNotes, which includes news-like corpora with a relatively standardized entity label system and coarse granularity; CLUENER, derived from multiple sources such as news, encyclopedias, and social media, featuring finer label granularity, broader coverage across multiple domains, and a high-coverage label system; and BosonNER, primarily based on social media platforms like Weibo, with a relatively flat but practical label system. These datasets differ in entity types, annotation styles, and domain backgrounds, making them suitable as typical examples for integrating multi-source heterogeneous NER corpora. Specific details about the datasets are shown in Table \ref{tab3}. The numbers in parentheses indicate the quantity of entities for each type.

\begin{table}[h]
\caption{Details of NER datasets}\label{tab3}%
\begin{tabular}{
    >{\centering\arraybackslash}m{0.11\textwidth} 
    >{\centering\arraybackslash}m{0.10\textwidth} 
    >{\centering\arraybackslash}m{0.10\textwidth} 
    >{\centering\raggedright\arraybackslash}m{0.55\textwidth}}
\toprule
Name & Size  & Domain &  Entity types(\#)\\
\midrule
OntoNotes & 900K(500K in Chinese Portion) & Mixed & PERSON(13506), EVENT(1208), CARDINAL(8703), ORG(10363),DATE(10029), NORP(3214), GPE(19221), LOC(2565), MONEY(1452), WORK\_OF\_ART(1012), TIME(1847), ORDINAL(1408), QUANTITY(1058), FAC(1514), PRODUCT(375), PERCENT(1009), LANGUAGE(345), LAW(312)\\
BosonNER & 2k & Social Media & Person(5141), location(4597), organization(2689), time(4250), company(2374), product(4122)\\
CLUENER & 12k+ & News & Person(4112), organization(3419), position(3477), company(3263), address(3193), game(2612),  government(2041), scene(1661), book(152), movie(1259)\\
\botrule
\end{tabular}
\end{table}

In the alternative scenario, the self-constructed small dataset FinReportNER in the financial domain is used. It consists of 823 annotated sentences and nine entity categories, which are \texttt{RATIO, TIME, NUM, FTERM, INDUSTRY, ORG, TREND, PRODUCT, EVENT}. It acts as a representative low-resource dataset to evaluate the generalization ability of our proposed label merging strategy in data-scarce situations (see Section \ref{subsec4_5}).

\textbf{Dataset partitioning:}
Each dataset follows its official train, validation, and test split. Only the training sets are used for model training, label similarity calculation, and merging. Validation sets are used to evaluate the latest model during the training epoch. Test sets are used to evaluate the impact of merging.

\textbf{Model Architecture and Training Parameters:} A BERT-based model, Chinese-BERT-wwm-ext\cite{60Cui2021ChineseBertWWM}, with CRF, is used as the NER model to evaluate the impact of different label merging results. Compared to encoder-decoder and decoder-only architectures, encoder-only models provide a good balance between performance and efficiency in sequence labeling tasks. Additionally, it utilizes the Whole Word Masking (WWM) mechanism, which enhances the modeling of Chinese word semantics by masking entire words during pre-training, rather than individual characters. This results in improved contextual representation, especially for longer or compound entities, and has demonstrated better performance on the Chinese NER task. We fine-tuned the model using AdamW optimizer\cite{61HE2025Adamw}, with a learning rate of 3e-5 for the BERT layer and 3e-2 for the CRF layer, over a maximum of 30 epochs. The batch size was 32, and the maximum sequence length was set to 150. All models are trained on an NVIDIA RTX 3080Ti graphics card. After each training round, the validation set is employed to assess the model, and the model with the highest F1 score is saved as the final version.

\subsection{Empirical Similarity and Merging path}\label{subsec4_2}

\begin{figure}[htbp]
  \centering
  \subfloat[CLUENER $\rightarrow$ BosonNER]{%
    \includegraphics[width=0.3\textwidth]{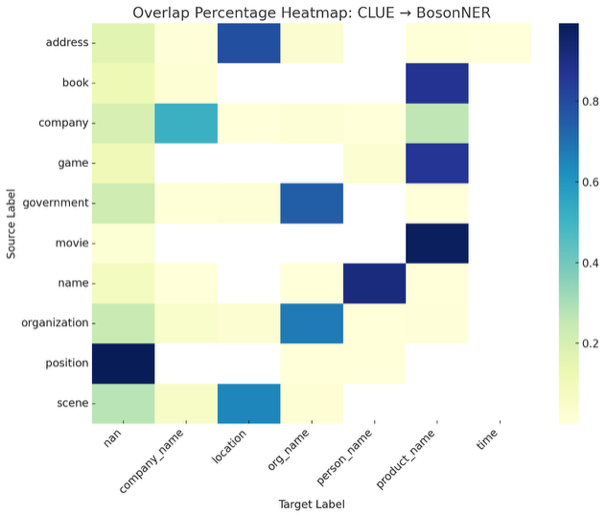}
  }
  \hfill
  \subfloat[BosonNER $\rightarrow$ CLUENER]{%
    \includegraphics[width=0.3\textwidth]{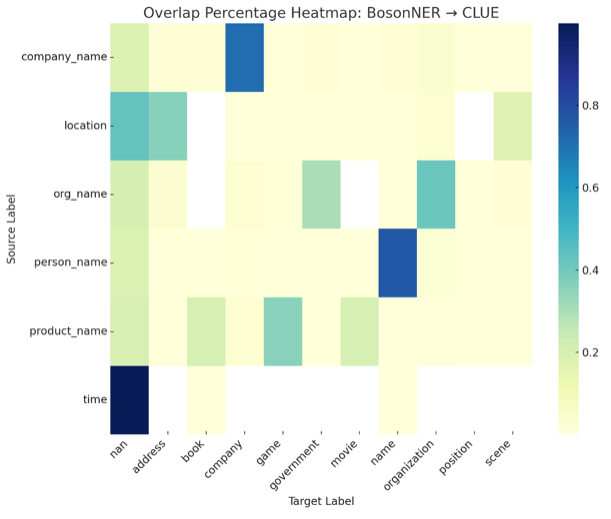}
  }
  \hfill
  \subfloat[CLUENER $\rightarrow$ OntoNotes]{%
    \includegraphics[width=0.3\textwidth]{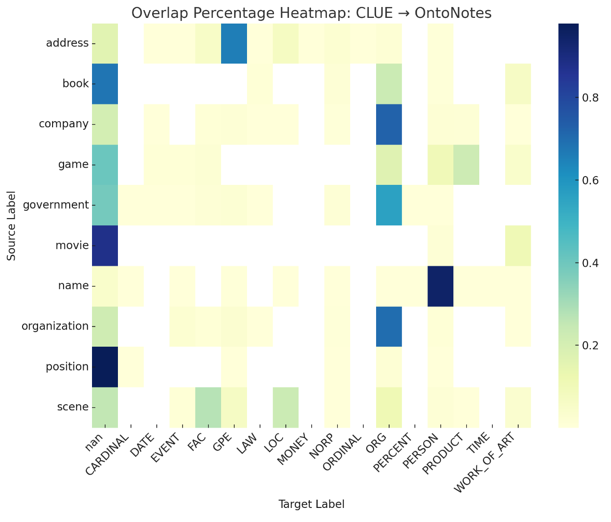}
  }

  \subfloat[OntoNotes $\rightarrow$ CLUENER]{%
    \includegraphics[width=0.3\textwidth]{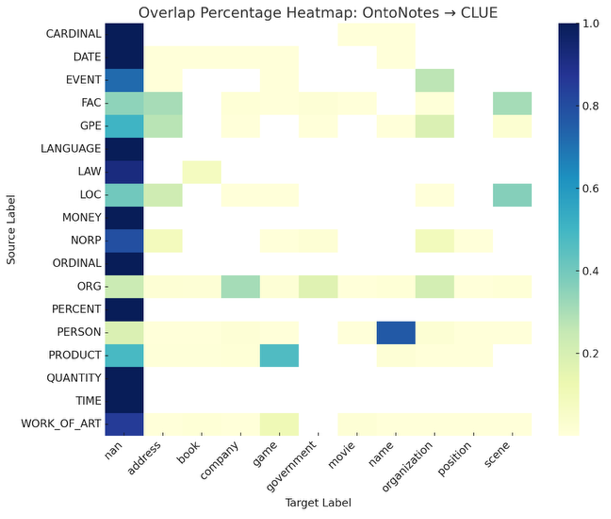}
  }
  \hfill
  \subfloat[BosonNER $\rightarrow$ OntoNotes]{%
    \includegraphics[width=0.3\textwidth]{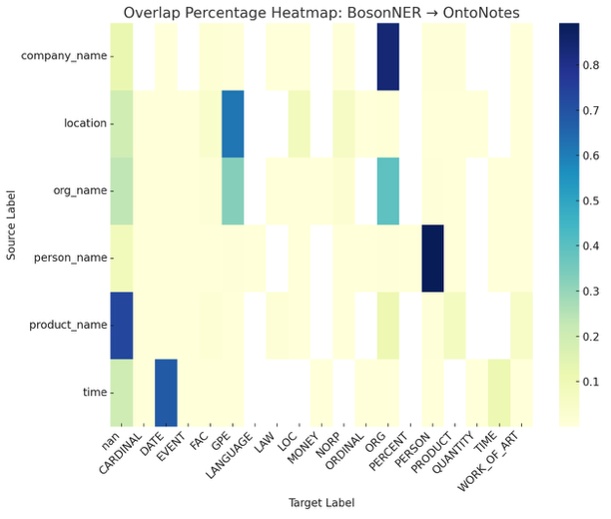}
  }
  \hfill
  \subfloat[OntoNotes $\rightarrow$ BosonNER]{%
    \includegraphics[width=0.3\textwidth]{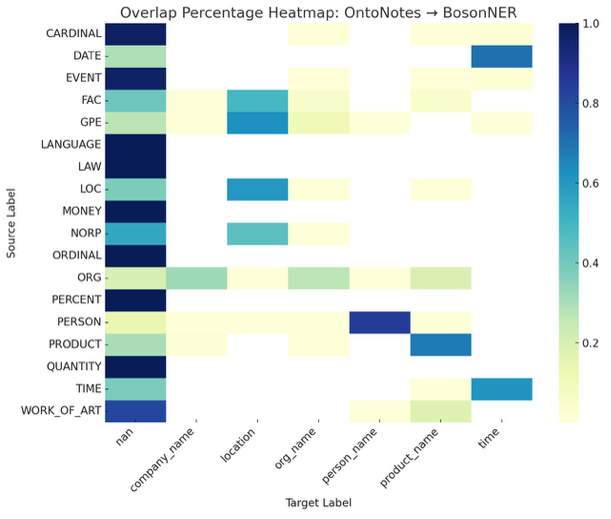}
  }
  \caption{Empirical similarity heatmaps across dataset pairs}
  \label{fig:empirical_sim_all}
\end{figure}

We calculated the empirical similarity score between each pair of datasets (CLUENER, BosonNER, and OntoNotes) to provide a quantitative basis for label alignment and merging paths selection. Therefore, we conducted six experiments, considering bidirectional calculation: CLUENER $\leftrightarrow$ BosonNER, CLUENER $\leftrightarrow$ OntoNotes, and BosonNER $\leftrightarrow$ OntoNotes. For example, in the experiment CLUENER $\rightarrow$ BosonNER, we trained a NER model on the training set of CLUENER, which treated as the source dataset $D_s$. Then this model was applied to the training set of BosonNER, treated as the target dataset $D_t$, to generate entity predictions. We collected the predicted entities labeled as $N_{\text{pred}}(L_s)$, and compared them with the gold-standard entities annotated as $N_{\text{true}}(L_t)$ in BosonNER. The empirical similarity $S_{\text{empirical}}(L_s, L_t)$ is calculated as the proportion of entities predicted as $L_s$ that are annotated as $L_t$ in the target dataset. All resulting empirical similarity matrices are visualized as heatmaps in Figure \ref{fig:empirical_sim_all} for intuitive comparison.

The results show a significant label mapping advantage between CLUENER and BosonNER. For example, the fine-grained labels such as \texttt{book}, \texttt{movie}, and \texttt{game} in CLUENER can be merged into \texttt{product\_name} in BosonNER. At the same time, \texttt{organization} and \texttt{government} highly overlap with \texttt{org\_name}, indicating that the label granularity of CLUE is finer and BosonNER is more coarse-grained. Based on the sum of empirical similarity scores across all label pairs, the CLUENER $\rightarrow$ BosonNER direction achieves the highest aggregated value among all dataset pairs. Therefore, the initial merging path is selected as CLUENER $\rightarrow$ BosonNER, forming an intermediate dataset (denoted as BosonM), which retains a relatively complete fine-grained label system.
Subsequently, the empirical similarity between BosonM and OntoNotes is analyzed. It is found that labels such as \texttt{ORG}, \texttt{GPE}, and \texttt{PRODUCT} in OntoNotes have a high matching degree with the labels in BosonM, making it suitable for further merging. According to the combined empirical similarity scores, BosonM $\rightarrow$ OntoNotes is chosen as the second merging path. This leads to the creation of a comprehensive, unified NER dataset that integrates detailed label hierarchies with broad cross-domain applicability.

\subsection{Semantics Similarity}\label{subsec4_3}

To further analyze the semantic similarity of named entity labels across datasets and facilitate label alignment, we conducted two experiments. These experiments used the merging path selected in Section \ref{subsec4_2} to compute and visualize a semantic similarity matrix and a 2D embedding graph. The process employed the Chinese-BERT-wwm-ext model for entity representation, combined with mean pooling, centralization, and normalization. Ultimately, cosine similarity was used to measure the semantic similarity between labels.

\begin{figure}
    \centering
    \includegraphics[width=0.75\linewidth]{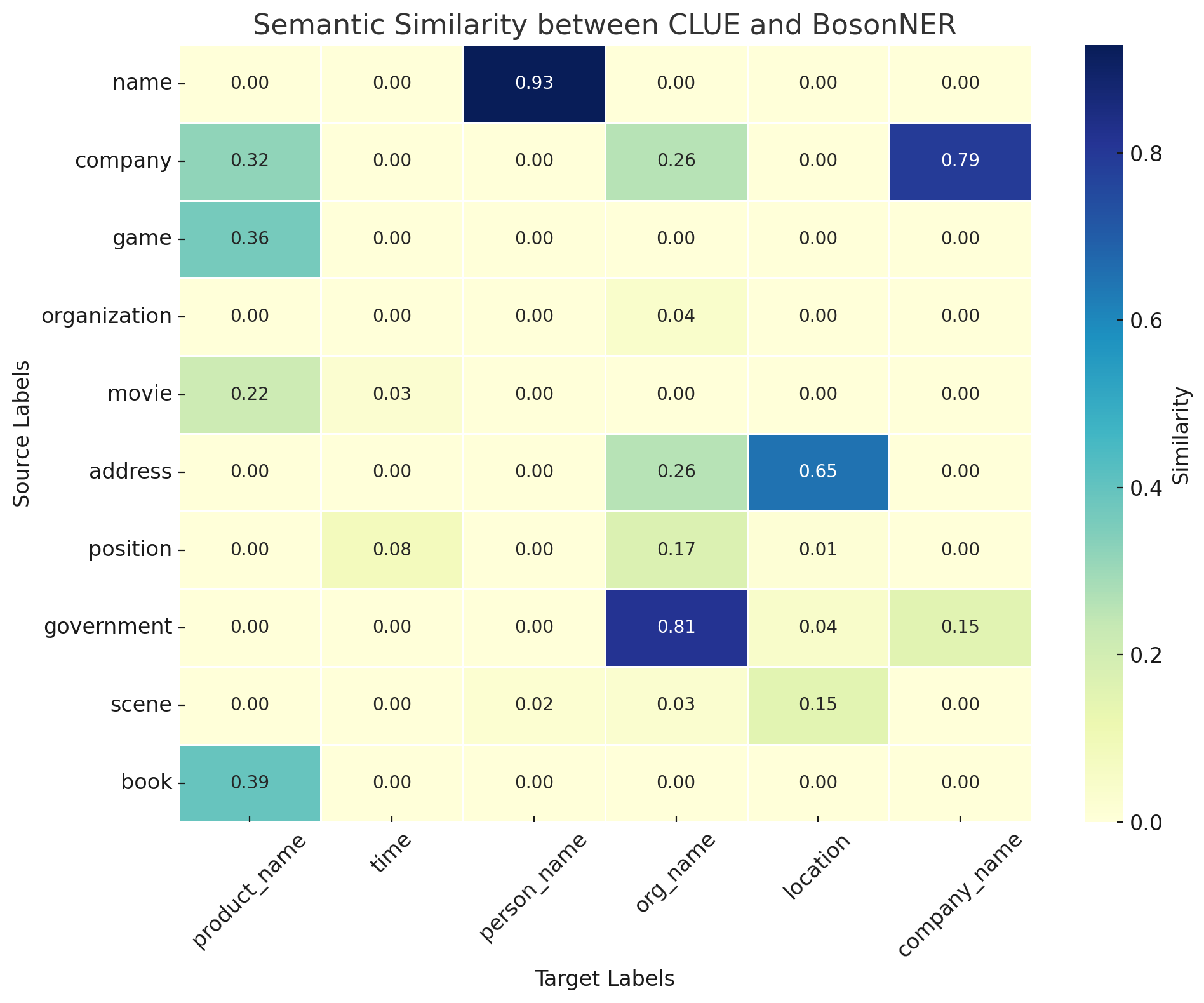}
    \caption{Semantic similarity between CLUE and BosonNER}
    \label{fig:semantic_similarity_clue_boson}
\end{figure}

In the first experiment, we analyzed semantic similarity between CLUENER and BosonNER labels. First, we extracted all the entities corresponding to each label from each dataset and obtained the context embedding vectors of the entities through the Chinese-BERT-wwm-ext model. Then, mean vectors were computed for each label after centralization and normalization. Finally, we build a semantic similarity matrix by calculating the cosine similarity between the mean vectors of label pairs. By analyzing Figure \ref{fig:semantic_similarity_clue_boson}, it shows that CLUENER’s fine-grained labels, such as \texttt{book}, \texttt{movie}, and \texttt{game}, are semantically close to \texttt{product\_name} in BosonNER. This reflects the ability of \texttt{product\_name} in BosonNER to act as a semantically inclusive category. Furthermore, \texttt{government} aligns closely with \texttt{org\_name} (0.81), while \texttt{organization} (0.04) looks like no relation, indicating that CLUENER’s \texttt{government} tends to converge semantically in BosonNER’s organizational category. Similarly, \texttt{address} in CLUENER matches well with \texttt{location} (0.65) in BosonNER, indicating overlap in spatial references. Notably, \texttt{person\_name} in BosonNER displays a very high similarity with \texttt{name} (0.93) in CLUENER. They can be regarded as the same category. Overall, the heatmap shows that the majority of BosonNER labels are semantically close to labels in CLUENER. This indicates a very high possibility for merging.

\begin{figure}
    \centering
    \includegraphics[width=0.75\linewidth]{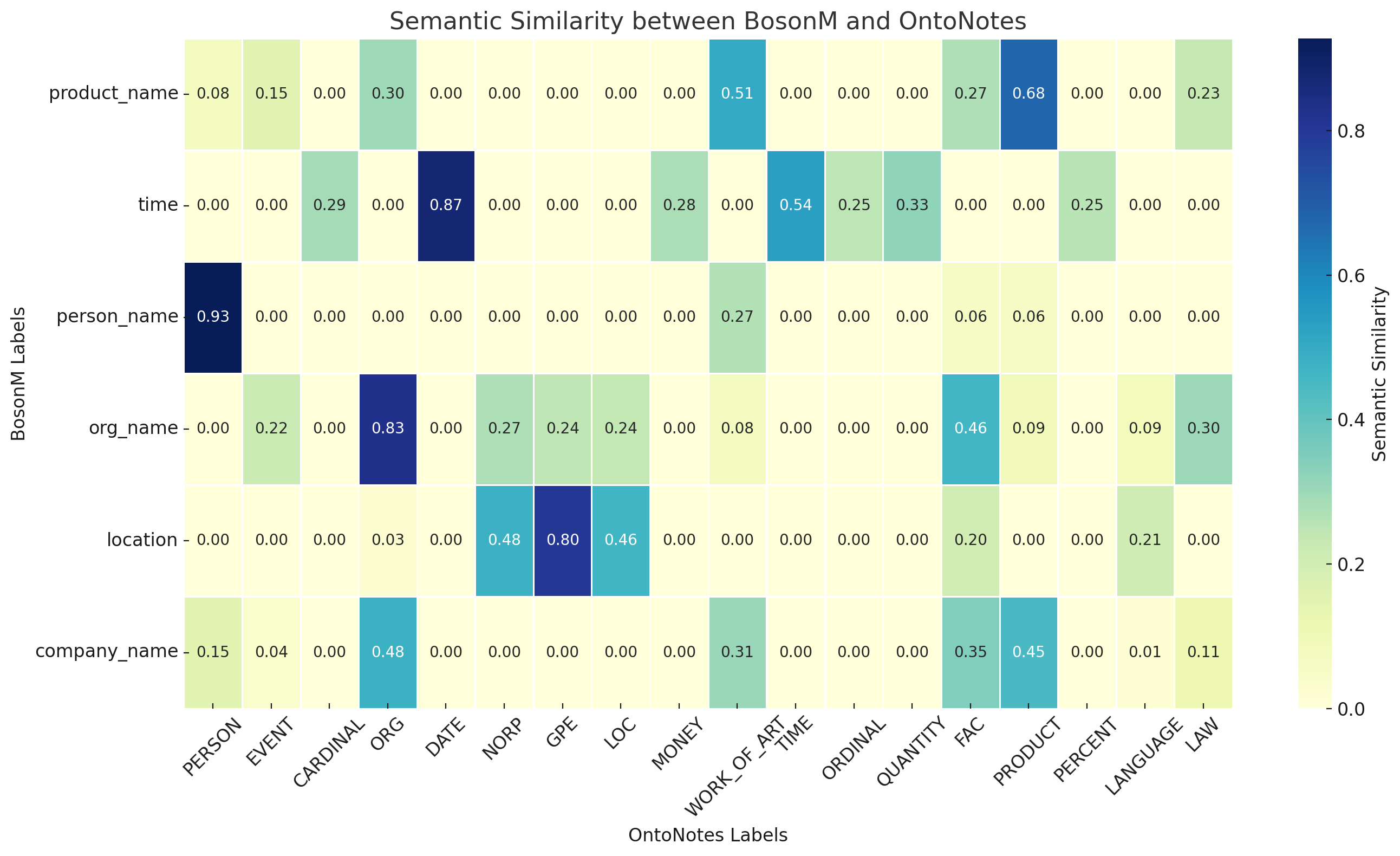}
    \caption{Semantic similarity between BosonM and Ontonotes}
    \label{fig:semantic_similarity_bosonm_onto}
\end{figure}

In the second experiment, we took the merged intermediate dataset BosonM as the new source dataset and conducted a semantic similarity analysis with OntoNotes. Since the labels in BosonM are new labels after fusion, their semantic representations need to be re-extracted and compared with the labels in OntoNotes. As shown in Figure~\ref{fig:semantic_similarity_bosonm_onto}, \texttt{location} in BosonM shows strong similarity with \texttt{GPE} (0.80), \texttt{LOC} (0.46), and \texttt{NORP} (0.48), indicating that this label covers geopolitical regions, general places, and demographic groups. Similarly, \texttt{org\_name} closely aligns with \texttt{ORG} (0.83), confirming their shared focus on organizations. Additionally, the label \texttt{org\_name} in BosonM aligns well with \texttt{ORG} (0.83), confirming their shared focus on organizational entities. Interestingly, \texttt{company\_name} in BosonM shows moderate semantic similarity with both \texttt{ORG} (0.48) and \texttt{PRODUCT} (0.45), reflecting its relevance across business and product contexts. Similarly, \texttt{product\_name} in BosonM correlates strongly with \texttt{PRODUCT} (0.68) and moderately with \texttt{WORK\_OF\_ART} (0.51), supporting its coverage of both normal goods and cultural products. \texttt{time} in BosonM maps effectively to \texttt{DATE} (0.87) and \texttt{TIME} (0.54), indicating that it tends to mark more date-related entities. Likewise, \texttt{person\_name} in BosonM demonstrates high similarity with \texttt{PERSON} (0.93) in OntoNotes, reinforcing its robustness in personal entity alignment. Overall, the semantic similarity matrix in Figure \ref{fig:semantic_similarity_bosonm_onto} confirms that BosonM can be effectively mapped onto OntoNotes, providing support for continued label merging and unified training.

\begin{figure}
    \centering
    \includegraphics[width=0.5\linewidth]{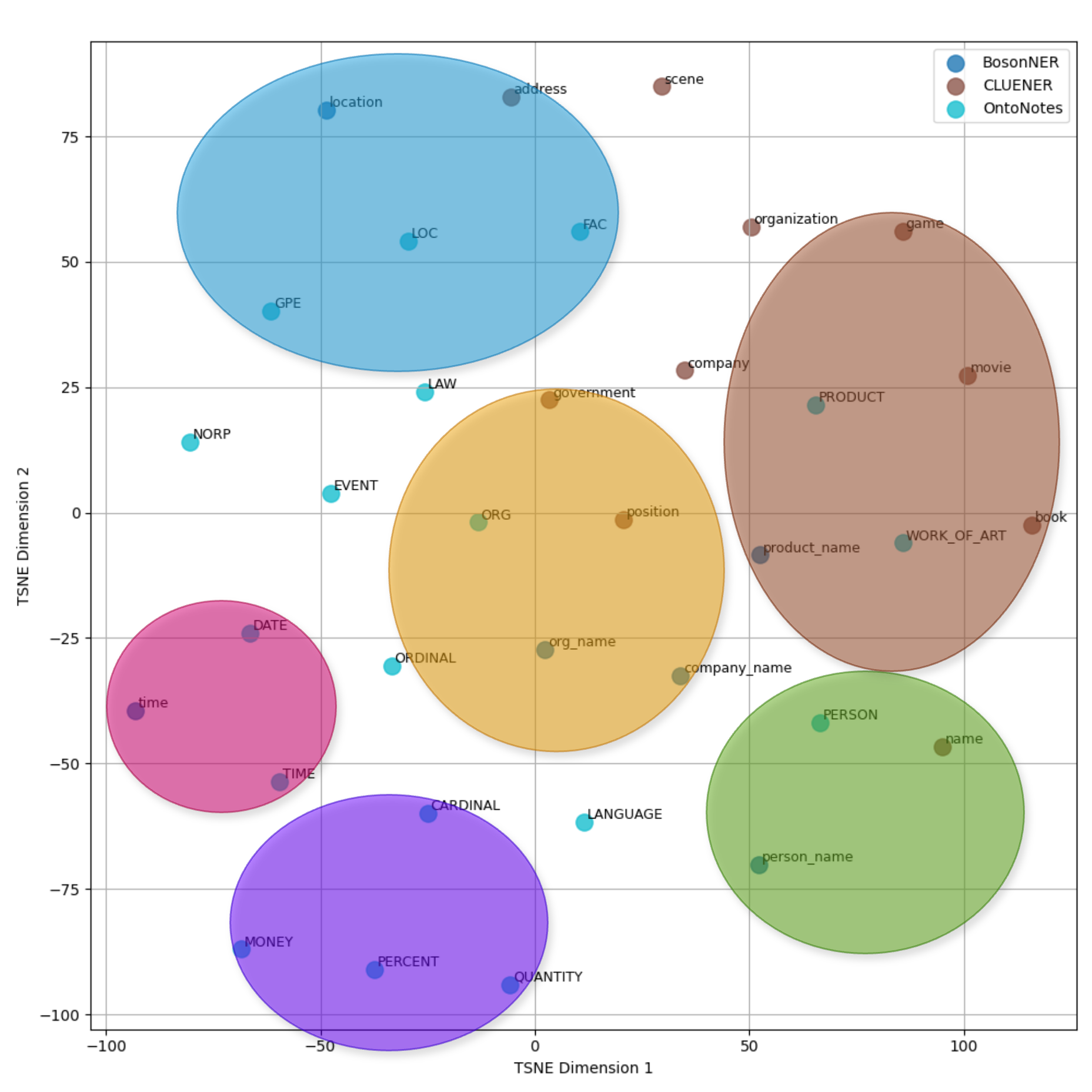}
    \caption{Semantic distribution of all labels from the original datasets (t-SNE)}
    \label{fig:2d_tsne}
\end{figure}

To further verify the overall semantic relationships among entity labels across datasets, we visualized the average embedding vectors of all labels using t-distributed Stochastic Neighbor Embedding (t-SNE)\cite{62maaten2008visualizing}, as shown in Figure \ref{fig:2d_tsne}. The resulting distribution clearly shows several clusters. Specifically, we observe that labels indeed form tight clusters across the three datasets:

\begin{itemize}
    \item Person names cluster (\texttt{PERSON}, \texttt{person\_name}, \texttt{name}) coincides in the bottom-right quadrant, confirming that these variants share nearly identical BERT representations and can be merged.
    \item The organizations cluster (\texttt{ORG}, \texttt{company\_name}, \texttt{org\_name}) occupies an adjacent region, indicating strong semantic overlap for ``organization/company'' entities.
    \item The locations cluster(\texttt{LOC}, \texttt{GPE}, \texttt{location}, \texttt{FAC}, \texttt{address}) is in the upper-left area, supporting their unification under a single ``location'' label.
    \item The product-related cluster (\texttt{PRODUCT}, \texttt{product\_name}, \texttt{book}, \texttt{movie}, \texttt{game}) gathers in the right-central region, reflecting that CLUENER’s fine-grained categories (\texttt{book/movie/game}) map closely onto the more general ``product'' concept.
    \item The numeric cluster (\texttt{QUANTITY}, \texttt{MONEY}, \texttt{PERCENT}) forms a distinct group in the lower-left, suggesting they can be consolidated into one ``numeric'' category or partitioned at a finer granularity if desired.
\end{itemize}

One notable exception is \texttt{position} in CLUENER, which appears embedded within the cluster of organizations. This suggests that \texttt{position} may co-occur with organizations frequently, leading to semantic overlap. In contrast, \texttt{organization} and \texttt{company} in CLUENER, which are theoretically expected to align closely with \texttt{ORG}, are instead located farther away in the upper-right region. This observation highlights that semantic similarity alone may not be sufficient to determine whether two labels should be merged, as it can be influenced by contextual noise or annotation inconsistencies. Therefore, incorporating empirical similarity based on actual prediction behavior is essential to ensure robust and reliable label alignment.

\subsection{ESNERA}\label{subsec4_4}

The experiment assesses the effect of label merging based on the comprehensive similarity $S_{merge}$, as defined in Section 3.1, which conducts interpolation on empirical and semantic similarity. The merging process follows the unidirectional similarity greedy merging strategy introduced in Section \ref{subsec3_2}, and the grid search optimization approach in Section \ref{subsec3_3}. The merging is conducted in two sequential stages based on the process mentioned above: First, CLUENER is merged into BosonNER, resulting in an intermediate dataset referred to as BosonM. Then, BosonM is further merged into OntoNotes to complete the label alignment process, resulting in a large-scale NER dataset. At each stage, we conducted grid search over the parameter space $\lambda\in{0.3,\ 0.4,\ 0.5,\ 0.6,\ 0.7}$ and $\tau\in{0.1,\ 0.2,\ 0.3,\ 0.4,\ 0.5,\ 0.6,\ 0.7,\ 0.8,\ 0.9}$, aiming to maximize the number of merged labels while constraining the drop in NER F1-score to be no more than 2\% compared to the baseline. The experiment results are summarized in Table \ref{tab4}.

\begin{table}
    \centering
    \begin{tabular}{>{\centering\arraybackslash}m{0.22\textwidth} 
                    >{\centering\arraybackslash}m{0.15\textwidth} 
                    >{\centering\arraybackslash}m{0.15\textwidth} 
                    >{\centering\arraybackslash}m{0.15\textwidth} 
                    >{\centering\arraybackslash}m{0.15\textwidth}}
    \toprule
        Method & \#Merged Label & Micro-F1 Score & $\Delta$ vs. Baseline 1 & $\Delta$ vs. Baseline 2 \\
    \midrule
        Baseline1 w/o Merging & N/A & 0.80 & N/A & N/A \\
        Baseline2 & 11 & 0.79 & -0.01 & N/A \\
        Proposed Method ($\lambda=0.3/0.4, \tau=0.4$) & 15 & 0.79 & -0.01 & 0 \\
        Proposed Method ($\lambda=0.5/0.6, \tau=0.3$) & 15 & 0.79 & -0.01 & 0 \\
    \botrule
    \end{tabular}
    \caption{Results of Label Merging on CLUE, BosonNER, and OntoNotes}
    \label{tab4}
\end{table}

Table \ref{tab4} showcases the experimental results of the performance comparison among different methods on the combination of CLUE, BosonNER, and OntoNotes. The overall results reveal that the micro-averaged F1 score of the proposed method is 0.79, which is the same as that of Baseline 2 (manual merging), but slightly lower than 0.80 of Baseline 1 (independently trained by OntoNotes). The proposed method merged 15 labels under both parameter settings (($\lambda$=0.3/0.4,$\tau$=0.4) and ($\lambda$=0.5/0.6,$\tau$=0.3)), outperforming the 11 labels of Baseline 2. In comparison with Baseline 2, the proposed method merged more labels while maintaining the same Micro-F1 score (0.79), indicating its superiority in label coverage; Baseline 1, without label merging, avoided semantic bias and had a slightly higher Micro-F1 score, but could not achieve label integration across datasets.

\begin{table}
    \centering
    \begin{tabular}{>{\centering\arraybackslash}m{0.10\textwidth} 
                    >{\centering\arraybackslash}m{0.10\textwidth} 
                    >{\centering\arraybackslash}m{0.10\textwidth} 
                    >{\centering\arraybackslash}m{0.10\textwidth} 
                    >{\centering\arraybackslash}m{0.10\textwidth}
                    >{\centering\arraybackslash}m{0.30\textwidth}}
    \toprule
    Label & Proposed Method F1	& Baseline 1 F1	& Difference & Support & Relation\&Merging Path \\
    \midrule
    PERSON & 0.92 & 0.92 & 0.00 & 1261 & Equivalence: name$\rightarrow$person\_name$\rightarrow$PERSON \\
    MONEY & 0.91 & 0.91 & 0.00 & 156 & Disjointness \\
    PERCENT & 0.86 & 0.87 & -0.01 & 177 & Disjointness \\
    GPE & 0.84 & 0.85 & -0.01 & 1778 & Subset: address, scene$\rightarrow$location$\rightarrow$GPE \\
    DATE & 0.82 & 0.82 & 0.00 & 976 & Equivalence: time$\rightarrow$DATE \\
    ORDINAL & 0.81 & 0.83 & -0.02 & 126 & Disjointness \\
    ORG & 0.78 & 0.79 & -0.01 & 1105 & Subset; company$\rightarrow$compant\_name$\rightarrow$ORG, government, organization$\rightarrow$org\_name$\rightarrow$ORG \\
    EVENT & 0.74 & 0.66 & +0.08 & 100 & Disjointness \\
    LOC & 0.71 & 0.72 & -0.01 & 268 & Disjointness \\
    CARDINAL & 0.66 & 0.67 & -0.01 & 742 & Disjointness \\
    NORP & 0.63 & 0.64 & -0.01 & 245 & Disjointness \\
    QUANTITY & 0.63 & 0.60 & +0.03 & 135 & Disjointness \\
    TIME & 0.67 & 0.67 & 0.00 & 160 & Disjointness \\
    LANGUAGE & 0.61 & 0.70 & -0.09 & 8 & Disjointness \\
    WORK \_OF\_ART & 0.56 & 0.55 & +0.01 & 63 & Disjointness \\
    FAC & 0.56 & 0.61 & -0.05 & 155 & Disjointness \\
    LAW & 0.53 & 0.63 & -0.10 & 17 & Disjointness \\
    PRODUCT & 0.34 & 0.58 & -0.24 & 35 & Partial Overlap: book, movie, game $\rightarrow$ product\_name  $\rightarrow$ PRODUCT  \\
    \midrule
    Micro-F1 & 0.79 & 0.80 & -0.01 & 7507 & N/A \\
    \botrule
    \end{tabular}
    \caption{Label-level comparison between the proposed method and Baseline 1}
    \label{tab:label_s1_f1_comparison}
\end{table}

Building on the overall Micro-F1 analysis, Table \ref{tab:label_s1_f1_comparison} shows a performance comparison at the label level between the proposed method ($\lambda = 0.4$, $\tau = 0.4$) and Baseline~1. The results indicate that while the proposed approach remains competitive, some labels experience nuanced changes due to label merging. For example, labels such as \texttt{PERSON} (F1: 0.92), \texttt{MONEY} (F1: 0.91), and \texttt{DATE} (F1: 0.82) maintain identical performance across both setups, indicating that their mappings (e.g., \texttt{name} $\rightarrow$ \texttt{person\_name} $\rightarrow$ \texttt{PERSON}) keep semantic consistency. \texttt{ORG} (F1: 0.78 vs. 0.79) stays strong despite combining \texttt{company}, \texttt{government}, and \texttt{organization}. In contrast, \texttt{PRODUCT} experiences a significant decline (F1: 0.34 vs. 0.58), likely due to semantic drift introduced by merging fine-grained categories like \texttt{book}, \texttt{movie}, and \texttt{game} into a broader label. Some labels, such as \texttt{EVENT} (+0.08) and \texttt{QUANTITY} (+0.03), benefit from additional training data, while others with limited support, like \texttt{LANGUAGE} and \texttt{LAW}, show decreased performance. 
Overall, the experimental results verify that the proposed method maintains competitive NER performance while expanding label coverage.

\subsection{Alternative Scenario with Small Dataset}\label{subsec4_5}

We annotated a small financial dataset, FinReportNER, to assess ESNERA's effectiveness in a resource-limited setting. In this case, we used the large dataset mentioned in Section \ref{tab:result_finreport} as the source, treating FinReportNER as the target. Following the same merging method—calculating empirical and semantic similarity and combining them with weights—we employed ESNERA to align and expand FinReportNER's label space. Evaluation was conducted on FinReportNER's dedicated test set, with the results summarized in Table \ref{tab:result_finreport}.

\begin{table}
    \centering
    \begin{tabular}{>{\centering\arraybackslash}m{0.20\textwidth} 
                    >{\centering\arraybackslash}m{0.15\textwidth} 
                    >{\centering\arraybackslash}m{0.15\textwidth} 
                    >{\centering\arraybackslash}m{0.15\textwidth} 
                    >{\centering\arraybackslash}m{0.15\textwidth}}
    \toprule
        Method & \#Merged Label & Micro-F1 Score & $\Delta$ vs. Baseline 1 & $\Delta$ vs. Baseline 2 \\
    \midrule
        Baseline1 w/o Merging & N/A & 0.74 & N/A & N/A \\
        Baseline2 Manual Merging& 6 & 0.77 & +0.03 & N/A \\
        Proposed Method ($\lambda=0.3/0.4, \tau=0.4$) & 5 & 0.77 & +0.03 & 0 \\
        Proposed Method ($\lambda=0.5/0.6, \tau=0.3$) & 5 & 0.77 & +0.03 & 0 \\
    \botrule
    \end{tabular}
    \caption{Results of Label Merging on FinReportNER}
    \label{tab:result_finreport}
\end{table}

Table~\ref{tab:result_finreport} reports the NER performance on FinReportNER under different merging strategies. The proposed method achieved a Micro-F1 score of 0.77, outperforming Baseline 1 (without merging) by 0.03, and the same with Baseline 2 (manual merging). While Baseline 2 merged six labels, our method merged five labels. The improvement over Baseline 1 indicates that label integration (e.g., merging \texttt{company} and \texttt{government} into \texttt{ORG}) effectively enhances recognition in the financial domain. The reason for the one less merging compared to Baseline 2 may be due to the insufficient sample size of the unmerged \texttt{EVENT} label in the FinReportNER dataset, which has only 10 samples. Importantly, the parameter settings obtained through grid search here match those in Section \ref{subsec4_4}, indicating that hyperparameters $\lambda$ and $\tau$ may have certain cross-dataset generalization capabilities. That is to say,  grid search can be skipped in other scenarios. 

\begin{table}[htbp]
    \centering
    \begin{tabular}{>{\centering\arraybackslash}m{0.14\textwidth}
                    >{\centering\arraybackslash}m{0.12\textwidth}
                    >{\centering\arraybackslash}m{0.12\textwidth}
                    >{\centering\arraybackslash}m{0.12\textwidth}
                    >{\centering\arraybackslash}m{0.10\textwidth}
                    >{\arraybackslash}m{0.30\textwidth}}
    \toprule
        Label & Proposed Path F1 & Baseline 1 F1 & Difference & Support & Relation \& Merge Path\\
    \midrule
        RATIO     & 1.00 & 1.00 & 0.00 & 26  & Equivalence: PERCENT$\rightarrow$RATIO \\
        TIME      & 0.91 & 0.88 & +0.03 & 47  & Subset: DATE$\rightarrow$TIME \\
        NUM       & 0.90 & 0.96 & -0.06 & 39  & Equivalence: MONEY$\rightarrow$NUM \\
        FTERM     & 0.82 & 0.84 & -0.02 & 38  & Disjointness \\
        INDUSTRY  & 0.82 & 0.80 & +0.02 & 187 & Disjointness \\
        ORG       & 0.87 & 0.78 & +0.09 & 26  & Subset: ORG$\rightarrow$ORG \\
        TREND     & 0.75 & 0.70 & +0.05 & 99  & Disjointness \\
        PRODUCT   & 0.59 & 0.52 & +0.07 & 119 & Subset: PRODUCT$\rightarrow$PRODUCT \\
        EVENT     & 0.24 & 0.31 & -0.07 & 10  & Disjointness \\
    \midrule
        Micro-F1  & 0.77 & 0.74 & +0.03 & 591 & N/A \\
    \botrule
    \end{tabular}
    \caption{Label-Level Evaluation on FinReportNER ($\lambda = 0.4$, $\tau = 0.4$)}
    \label{tab:finreport_label_eval}
\end{table}

To further evaluate the effectiveness of the proposed method, Table \ref{tab:finreport_label_eval} shows a comparison of label-level F1 scores between the proposed method and Baseline 1 on the FinReportNER test set. On core entity types, this method shows strong stability, such as \texttt{RATIO}, \texttt{TIME}, and \texttt{FTERM}. The \texttt{ORG} label has seen a significant improvement from 0.78 to 0.87, mainly due to merging semantically related labels such as \texttt{company} and \texttt{government}. \texttt{TREND} and \texttt{INDUSTRY} have also seen varying degrees of improvement, indicating that label diversity helps enhance performance. The performance of \texttt{PRODUCT} improved 0.07 after merging it with tags like \texttt{book} and \texttt{movie}. It is worth mentioning that the performance of the \texttt{NUM} slightly declined from 0.90 to 0.96, possibly due to the omission of some entities during the pseudo-labeling process. As mentioned in Section \ref{subsec3_4}, the source datasets (CLUENER and BosonNER) did not include annotations for amount-type entities. This incompleteness in annotation reduced the quality of the training signal, thereby limiting the performance improvement of the NUM category. The performance of the \texttt{EVENT} decreased from 0.24 to 0.31, likely because of its very small sample size (only 10 support), which makes it highly sensitive to merge errors. 

In summary, the proposed approach not only enhances label coverage but also preserves or improves the recognition performance of most labels. However, for labels that are limited by the number of samples or have merged deviations and noises, such as EVENT and NUM, further optimization remains necessary.

\section{Ablation Experiments}\label{sec5}

To verify the individual contributions of each component in ESNERA, the following ablation experiments were designed using the merged dataset from Section \ref{subsec4_4}: (1) \textbf{w/o Semantic Similarity}: Using only empirical similarity: Set $\lambda$=1 and disregard semantic similarity. (2)\textbf{w/o Empirical Similarity}: Using only semantic similarity: Set $\lambda$=0 and disregard empirical similarity.

\begin{table}
    \centering
    \begin{tabular}{>{\centering\arraybackslash}m{0.20\textwidth} 
                    >{\centering\arraybackslash}m{0.20\textwidth} 
                    >{\centering\arraybackslash}m{0.20\textwidth} 
                    >{\centering\arraybackslash}m{0.20\textwidth}}
    \toprule
    \textbf{Method} & \textbf{Micro-F1 Score} & \textbf{\#Merged Labels} & \textbf{$\Delta$ Merged Labels vs. Full} \\
    \midrule
    ESNERA (Full Model) & 0.79 & 15 &  N/A \\
    ESNERA w/o Semantic Similarity ($\lambda=1$) & 0.78  & 13 & –2 \\
    ESNERA w/o Empirical Similarity ($\lambda=0$) & 0.79 & 13 & –2\\
    \botrule
    \end{tabular}
    \caption{Ablation Results}
    \label{tab:ablation_main}
\end{table}

The experimental results are presented in Table \ref{tab:ablation_main}. The comprehensive model merged 15 labels, achieving an F1 score of 0.79. Upon the removal of semantic similarity, the number of merged labels diminished to 11, and paths such as \texttt{movie}$\rightarrow$\texttt{product\_name} and \texttt{company\_name}$\rightarrow$\texttt{ORG} were not successfully merged. When empirical similarity was excluded, the number of merged labels was 13, and paths \texttt{company}$\rightarrow$\texttt{company\_name} and \texttt{product\_name}$\rightarrow$\texttt{PRODUCT} were not captured. 

Although the F1 scores across all settings showed slight variation, due to the performance constraints discussed in Section \ref{subsubsec3_3_2}, the fluctuations in the number of merged labels highlight the different roles each module plays in label alignment. The findings indicate that combining empirical similarity and semantic similarity is crucial for achieving comprehensive and scalable label merging.

\section{Conclusion and Future Work}\label{sec6}
This study introduces an extensible label alignment method, ESNERA, aimed at unifying multiple source NER datasets by calculating label similarity. It employs a greedy pairwise merging strategy, which improves label coverage while maintaining model performance stability as much as possible. The core of ESNERA involves thoroughly calculating both empirical and semantic similarities between labels and automatically choosing the optimal merging parameters via a grid search mechanism to maximize label merges with less than 2\% performance loss in NER. 
Experiments across general and domain-specific datasets show that ESNERA can merge a large number of labels while maintaining NER performance and identifying their possible relations. Compared to separate training, the unified dataset improves label coverage and recognition of complex entity types. Ablation studies verify the necessity of each module, emphasizing the importance of integrating empirical and semantic signals for strong label alignment. Despite achieving certain results, ESNERA still has room for improvement. Future work will focus on: 1)The poor performance of some labels indicates challenges in cross-domain label alignment. Future research could explore hierarchical label modeling to determine more specific label relations. 2) Extending this method to multilingual scenarios and verifying its applicability in cross-language NER tasks are also important future directions. 

\section*{Declarations}

\subsection*{Competing Interests}
The authors declare no competing interests.

\subsection*{Data Availability Statement}
The datasets CLUENER2020, BosonNER, OntoNotes 5.0, and FinReportNER used in this study are publicly available from their respective official sources\footnote{\url{https://github.com/CLUEbenchmark/CLUENER2020}}\footnote{\url{https://catalog.ldc.upenn.edu/LDC2013T19}}\footnote{\url{https://github.com/InsaneLife/ChineseNLPCorpus/tree/master/NER/boson/}}. The dataset FinReportNER will be made available on reasonable request.

\makeatletter
\renewcommand\@biblabel[1]{#1.}
\makeatother
\bibliography{sn-bibliography}

\end{document}